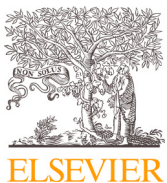
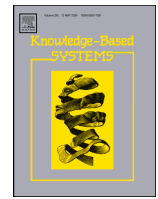

# CoSTI: Consistency models for (a faster) spatio-temporal imputation

Javier Solís-García [*], Belén Vega-Márquez, Juan A. Nepomuceno, Isabel A. Nepomuceno-Chamorro

*Dpto. Lenguajes y Sistemas Informáticos, University of Seville, Av. Reina Mercedes sn, Seville, 41012, Spain*

ABSTRACT

Multivariate Time Series Imputation (MTSI) is crucial for many applications, such as healthcare monitoring and traffic management, where incomplete data can compromise decision-making. Existing state-of-the-art methods, like Denoising Diffusion Probabilistic Models (DDPMs), achieve high imputation accuracy; however, they suffer from significant computational costs and are notably time-consuming due to their iterative nature. In this work, we propose CoSTI, an innovative adaptation of Consistency Models (CMs) for the MTSI domain. CoSTI employs Consistency Training to achieve comparable imputation quality to DDPMs while drastically reducing inference times, making it more suitable for real-time applications. We evaluate CoSTI across multiple datasets and missing data scenarios, demonstrating up to a 98 % reduction in imputation time with performance on par with diffusion-based models. This work bridges the gap between efficiency and accuracy in generative imputation tasks, providing a scalable solution for handling missing data in critical spatio-temporal systems. The code for this project can be found here: https://github.com/javiersgjavi/CoSTI.

## 1. Introduction

One of the most pervasive challenges when working with Multivariate Time Series (MTS) is handling missing values. This issue, known as MTSI, is critical because inaccurate imputations can distort the original data distribution [1]. Missing values arise from various situations, which differ across domains. For instance, they can result from sensor malfunctions in Internet of Things (IoT) applications [2,3], or in clinical settings [4], where patient data may be irregularly recorded or omitted entirely. In both cases, reliable and efficient imputation is essential to preserve downstream task performance.

Although traditional Machine Learning (ML) and recent Deep Learning (DL) methods have improved imputation accuracy [5,6], they often struggle to balance effectiveness and computational efficiency. Diffusion models, particularly Denoising Diffusion Probabilistic Models (DDPMs) [7], have achieved state-of-the-art performance. However, their iterative nature results in high inference costs, making them unsuitable for time-critical applications such as Intensive Care Unit (ICU) monitoring [1], traffic control [8,9], or energy systems [10].

To address these limitations, **Consistency Models** (CMs) [11] have recently been proposed. By distilling DDPMs into faster models or training them directly through Consistency Training, CMs offer a promising trade-off: significantly reduced inference times while maintaining competitive performance. Although CMs have shown promise in accelerating generative modeling tasks in domains such as computer vision, their application to time series, and specifically to imputation, has remained entirely unexplored. This work introduces the first adaptation of CMs to the MTSI problem.

In this work, we propose *Consistency models for (a faster) Spatio-Temporal Imputation* (CoSTI), an efficient and effective framework for MTSI based on Consistency Models, for which a graphical representation of the concept can be seen in Fig. 1. Our key contributions are as follows:

1. We introduce a novel application of Consistency Models to MTSI, and propose a tailored adaptation of Consistency Training for this domain.
2. We design a spatio-temporal architecture that integrates conditional information and curriculum-based training to improve robustness and accuracy.
3. We demonstrate through extensive experiments that CoSTI reduces inference time by up to 98 % compared to DDPM-based models, while achieving comparable accuracy across diverse datasets and missing-data scenarios.

The rest of this paper is structured as follows: Section 2 reviews related work. Section 3 provides the mathematical foundations of our approach. Section 4 introduces CoSTI. Section 5 describes the experimental setup and results. Finally, Section 6 concludes the paper. Additional implementation details and code can be found in Appendix A.

* Corresponding author.
*E-mail address:* jsolisg@us.es (J. Solís-García).

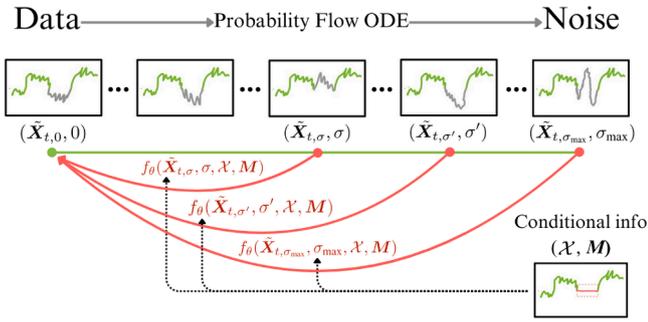

**Fig. 1.** The objective is to map missing values back to their origin by leveraging a consistency model with conditional information. Given a Probability Flow ODE (PF-ODE) that progressively adds noise to the missing values, the model learns to reconstruct points along the ODE trajectory, allowing accurate imputations without the need for a full reverse path.

## 2. Related works

*Multivariate time series imputation.* The challenge of MTSI has been extensively studied through a variety of approaches. Traditional methods, such as mean imputation, zero imputation, or linear trends [4], are simple to implement but often compromise data integrity by introducing bias. ML techniques, including k-nearest neighbors [12], matrix factorization [13], and scalable Gaussian processes [14], offer more adaptive and robust solutions. DL further transformed the field with Recurrent Neural Networks (RNNs), which excel at capturing sequential dependencies critical for imputation tasks [15,16]. Bidirectional RNNs (BiRNNs), such as BRITS [5], enhance this capability by processing sequences in both forward and reverse temporal directions.

Transformers, with their self-attention mechanisms, address limitations of RNNs such as error accumulation and can capture long-range dependencies in time series data [17,18]. Graph Neural Networks (GNNs) have also been widely adopted to capture spatial dependencies between variables. A common practice in recent MTSI works is to model spatial relationships via a fixed graph structure, where the adjacency matrix remains constant over time [6].

Recent efforts in tensor-based methods have explored the imputation problem from a high-dimensional perspective. For instance, low-rank tensor completion methods such as FDPTC [19], NT-DPTC [8], and HRST-LR [9] incorporate spatio-temporal regularizations or preserve specific dimensions to enhance imputation accuracy in traffic data.

Generative methods have shown significant potential for MTSI due to their ability to model complex data distributions. Early approaches such as Variational Autoencoders (VAEs) [20] and Generative Adversarial Networks (GANs) [21] advanced the field, although GANs often suffer from mode collapse [22]. The combination of generative models with GNNs has further improved the ability to capture multivariate dependencies [18], leading to increasingly robust MTSI pipelines.

*Diffusion models for MTSI.* DDPMs have emerged as a robust solution for MTSI, leveraging iterative denoising processes to reconstruct missing data with high accuracy. Conditioning mechanisms, as demonstrated by CSDI [17], are critical to their success, allowing DDPMs to outperform earlier generative methods. Conversely, unconditional DDPMs fail to deliver accurate imputations [23], underscoring the importance of carefully crafted conditional architectures. Recent advancements, such as PriSTI [18], have refined this approach by introducing specialized layers to enhance the generation of conditional information, achieving state-of-the-art results. Furthermore, the integration of S4 blocks into DDPMs has shown promise in improving temporal modeling efficiency [24].

Recent advancements have further demonstrated the flexibility of DDPM architectures. TIMBA [25], for instance, replaces Transformer-based temporal blocks with bidirectional Mamba blocks, achieving competitive performance and showcasing the effectiveness of structured attention mechanisms in diffusion models. This approach highlights how architectural adaptations can further enhance the robustness and efficiency of DDPMs for MTSI, reinforcing their status as a leading framework for accurate spatio-temporal imputation.

*Consistency models.* Consistency Models (CMs) are a recent advancement in generative modeling, closely related to DDPMs, but designed to optimize for single-step generation. This approach prioritizes faster inference, albeit with a potential trade-off in generation quality. CMs can be developed either through *Consistency Distillation* (CD), where a diffusion model is distilled into a single-step generator, or trained independently using *Consistency Training* (CT) [11].

Initially, CMs excelled in speed but struggled to match the quality of diffusion models. Over time, refined methodologies and implementations have significantly bridged this gap, enabling CMs to achieve performance comparable to state-of-the-art diffusion models. While CD has historically been the dominant training paradigm, it involves training a diffusion model as an intermediate step, increasing computational cost. Recent advancements have demonstrated that CT can surpass CD in performance while eliminating the need for an intermediary diffusion model, reducing both training time and resource requirements [26–28].

Despite their potential, CMs remain underexplored due to their relatively recent introduction. Current applications are largely confined to computer vision, where approaches such as Latent Consistency Models [29] have leveraged Latent Diffusion Models [30] to achieve faster generation speeds without compromising output quality. However, to the best of our knowledge, CMs have yet to be applied to time series data or MTSI, presenting a promising opportunity to exploit their rapid generation capabilities in this domain.

## 3. Background

### 3.1. Multivariate time series imputation

A MTS consists of recordings at different time steps, denoted by $t$, across multiple variables or channels, denoted by $N_t$. Formally, an MTS can be represented as $X_t \in \mathbb{R}^{N_t \times d}$, where each row $i$ contains the $d$-dimensional vector $x_t^i \in \mathbb{R}^d$, corresponding to the $i$-th variable at time $t$. Additionally, a binary mask matrix $M_t \in \{0,1\}^{N_t \times d}$ is used to indicate the presence of missing values: $M_t[i,j] = 0$ when the corresponding value in $X_t$ is missing, and $M_t[i,j] = 1$ when the value is observed.

The imputed time series generated by a model is represented as $\hat{X}_t$, while $\tilde{X}_t$ and $X_t$ denote the data with missing values and ground-truth imputation we aim to approximate, and $\mathcal{X}_t$ refers to the time series imputed using a linear interpolation technique.

Since our approach leverages graph-based models, following Cini et al. [6], we model the MTS as a sequence of graphs. At each time step $t$, a graph $\mathcal{G}_t = \langle \tilde{X}_t, \mathcal{A}_t \rangle$ is defined, where $\mathcal{A}_t$ is the adjacency matrix for that time step. In this work, we assume a constant graph topology over time; thus, $\mathcal{A}$ remains fixed for all $t$. This matrix is typically constructed once from domain knowledge (e.g., sensor layout) or computed via pairwise similarity (e.g., correlation), and reused throughout the sequence. This strategy is widely adopted in prior works and allows the model to capture spatial relationships efficiently without recomputing the structure at each time step [6,18].

### 3.2. Consistency models

CMs are built on the probability flow ordinary differential equation (PF-ODE), which describes the evolution of data corrupted by Gaussian noise. Starting from a data distribution $p_{data}(\mathbf{x})$, the diffusion process applies noise perturbations $\mathcal{N}(0, \sigma)$, transforming the data into pure noise. This process is expressed as [26]:

$$p_\sigma(\mathbf{x}) = \int p_{data}(\mathbf{y}) \mathcal{N}(\mathbf{x}|\mathbf{y}, \sigma^2 I) d\mathbf{y} \qquad (1)$$

The PF-ODE is given by [31]:

$$\frac{d\mathbf{x}}{d\sigma} = -\sigma \nabla_x \log p_\sigma(\mathbf{x}) \quad \sigma \in [\sigma_{\min}, \sigma_{\max}] \quad (2)$$

where $\nabla_x \log p_\sigma(\mathbf{x})$, referred to as the score function, measures the gradient of the log-likelihood of the data under $p_\sigma(\mathbf{x})$. Diffusion models are a type of *score-based generative model* because they estimate this score function at different noise levels during the corruption process [32]. For practical implementation, $\sigma_{\min}$ is set to a small value such that $p_{\sigma_{\min}}(\mathbf{x}) \approx p_{\text{data}}(\mathbf{x})$, and $\sigma_{\max}$ is large enough to approximate $p_{\sigma_{\max}}(\mathbf{x}) \sim \mathcal{N}(0, \sigma_{\max}^2 \mathbf{I})$. Following Karras et al. [31], we set $\sigma_{\min} = 0.002$ and $\sigma_{\max} = 80$.

The PF-ODE allows transitioning between noise levels. For example, $\mathbf{x}_{\sigma_{\min}}$ can be recovered from $\mathbf{x}_\sigma$ by solving the PF-ODE, defining a *consistency function* $f^* : (\mathbf{x}_\sigma, \sigma) \to \mathbf{x}_{\sigma_{\min}}$. This function must satisfy *self-consistency*: $f^*(\mathbf{x}_\sigma, \sigma) = f^*(\mathbf{x}_{\sigma'}, \sigma')$ for any $\sigma, \sigma' \in [\sigma_{\min}, \sigma_{\max}]$. Additionally, it must respect the *boundary condition* $f^*(\mathbf{x}, \sigma_{\min}) = \mathbf{x}$, where it acts as an identity function [11].

The consistency function is parameterized using a neural network called the *consistency model*, $F_\theta$, which is designed to approximate $f^*$ while adhering to the boundary condition. The parameterization used in this work, following Song et al. [11], is:

$$f_\theta(\mathbf{x}, \sigma) = c_{\text{skip}}(\sigma)\mathbf{x} + c_{\text{out}}(\sigma)F_\theta(\mathbf{x}, \sigma) \quad (3)$$

With differentiable functions $c_{\text{skip}}(\sigma)$ and $c_{\text{out}}(\sigma)$ such that $c_{\text{skip}}(\sigma_{\min}) = 1$ and $c_{\text{out}}(\sigma_{\min}) = 0$, defined as:

$$c_{\text{skip}}(\sigma) = \frac{\sigma_{\text{data}}^2}{(\sigma - \sigma_{\min})^2 + \sigma_{\text{data}}^2}, \quad c_{\text{out}}(\sigma) = \frac{\sigma_{\text{data}}^2 (\sigma - \sigma_{\min})}{\sqrt{\sigma_{\text{data}}^2 + \sigma^2}} \quad (4)$$

Training involves discretizing the PF-ODE into noise levels $\sigma_{\min} = \sigma_1 < \cdots < \sigma_N = \sigma_{\max}$, parameterized as $\sigma_i = \left(\sigma_{\min}^{1/\rho} + \frac{i-1}{N-1}\left(\sigma_{\max}^{1/\rho} - \sigma_{\min}^{1/\rho}\right)\right)^\rho$ with $\rho = 7$ [31].

The training objective minimizes the *consistency matching* loss:

$$\mathcal{L}^N(\theta, \theta^-) = \mathbb{E}[\lambda(\sigma_i) d(f_\theta(\mathbf{x}_{\sigma_{i+1}}, \sigma_{i+1}), f_{\theta^-}(\mathbf{x}_{\sigma_i}, \sigma_i))] \quad (5)$$

where $d(\mathbf{x}, \mathbf{y})$ is the Pseudo-Huber metric [33], $d(\mathbf{x}, \mathbf{y}) = \sqrt{\|\mathbf{x} - \mathbf{y}\|_2^2 + c^2} - c$, with $c = 0.00054\sqrt{d}$, and $\lambda(\sigma) = 1/(\sigma_{i+1} - \sigma_i)$. In this framework, $f_\theta$ is the *student network*, while $f_{\theta^-}$ is the *teacher network*. Both networks share the same parameters, but during training, $f_{\theta^-}$ executes its forward pass using a stopgrad operation, preventing gradient computation for this network and ensuring stability. Noise levels $i$ are sampled using a discrete Lognormal distribution parameterized as $i \sim p(i) \propto \text{erf}\left(\frac{\log(\sigma_{i+1} - P_{\text{mean}})}{\sqrt{2}P_{\text{std}}}\right) - \text{erf}\left(\frac{\log(\sigma_i - P_{\text{mean}})}{\sqrt{2}P_{\text{std}}}\right)$, where $P_{\text{mean}} = -1.1$ and $P_{\text{std}} = 2.0$ [26].

## 4. CoSTI

### 4.1. Consistency training for imputation

To train a Consistency Model for MTSI, we introduced several modifications to improve the results. Below, we detail the adjustments made and describe the final architecture used to implement the Consistency Model.

#### 4.1.1. Conditional information

Consistency models share many similarities with diffusion models, as they are heavily inspired by the underlying theoretical framework. Diffusion models have been successfully applied in the field of MTSI, often emphasizing the importance of incorporating enhanced conditional information to guide the model towards better imputations.

In light of these findings and building on the successes of diffusion-based models in this domain, we incorporated conditional information following a philosophy similar to Liu et al. [18]. Specifically, we included $\mathcal{X}_t$, $\mathcal{A}$ and $\mathbf{M}_t$. Consequently, our Consistency Model is parameterized as follows: $F_\theta : (\mathbf{X}_{t,\sigma_i}, \mathcal{X}_t, \mathcal{A}, \mathbf{M}_t, \sigma_i) \to \hat{\mathbf{X}}_t$

#### 4.1.2. Loss function and regularization

In MTSI, the task requires reconstructing information often entirely absent in the original dataset. To simulate this scenario during training, synthetic missing values are dynamically generated in each batch. The error is then computed solely at positions where ground-truth information exists. This process yields $\tilde{X}_t$, a data matrix with synthetic missing values, and $\tilde{M}_t$, an updated mask matrix reflecting these values. Incorporating these updates and the conditional information from Section 4.1.1, the loss function is defined as:

$$\mathcal{L}^N(\theta, \theta^-) = \mathbb{E}[\mathbf{M} \cdot \lambda(\sigma_i) d(f_\theta(\tilde{X}_{t,\sigma_{i+1}}, \mathcal{X}_t, \mathcal{A}, \tilde{M}_t, \sigma_{i+1}), \\ f_{\theta^-}(\tilde{X}_{t,\sigma_i}, \mathcal{X}_t, \mathcal{A}, \tilde{M}_t, \sigma_i))]. \quad (6)$$

Additionally, strong regularization is critical for CMs. We observed that increasing the dropout rate from 0.1 [25] to 0.2 improved performance, underscoring the importance of mitigating overfitting during training. This adjustment, along with synthetic missing values, enhances the model's robustness in handling diverse imputations.

#### 4.1.3. Optimization strategy and curriculum learning

Having described the loss function and regularization techniques, we now turn our attention to the optimization strategy and training dynamics, which are critical to ensuring convergence and stability in CoSTI. Optimization strategies in prior work have varied: Song et al. [11] used RAdam without schedulers or weight decay, while Geng et al. [27] incorporated learning rate decay in their experiments with RAdam and Adam. In our approach, we adopted the Scheduler-Free optimizer [34], specifically the AdamW variant, which eliminates the need for a learning rate scheduler while achieving state-of-the-art results. To ensure stable convergence, we introduced weight decay, further emphasizing the role of regularization in training Consistency Models.

For curriculum learning, we progressively increased $N$ during training to refine consistency across different $\sigma$ values. This strategy allows the model to start by comparing points farther apart along the PF-FLOW and gradually focus on smaller differences between closer $\sigma$ values. In this work, we used a linear scheduler, incrementing $N$ from $s_0 = 10$ to $s_1 = 200$, achieving a balance between simplicity and effective training. The function defining the evolution of $N$ according to the train step $k$ for our currirulum scheduler is defined as $N(k)$.

#### 4.1.4. Deterministic imputation

Like diffusion models, Consistency Models yield a probabilistic imputation by sampling from a learned distribution. Let $\{\hat{X}_t^{(i)}\}_{i=1}^N$ denote the $N$ independent imputations of the time series generated by the model through stochastic forward passes. To obtain a deterministic imputation that is robust to variability and outliers, we compute the element-wise median over these samples. Formally, the final deterministic imputation is defined as:

$$\hat{X}_{t,det} = \text{median}\left(\left\{\hat{X}_t^{(i)}\right\}_{i=1}^N\right) \quad (7)$$

This approach is grounded in the statistical robustness of the median, which minimizes the expected absolute deviation and provides a reliable point estimate that represents the distribution predicted by the model. In our experiments, we set $N = 100$, ensuring that $\hat{X}_{t,det}$ reliably approximates the true data distribution while mitigating the influence of extreme values.

### 4.2. Model architecture

With the training strategy in place, we now present the architecture of CoSTI, designed to effectively capture the spatio-temporal structure of multivariate time series data while leveraging the strengths of consistency models. Our design follows a U-Net [35] backbone with dual branches, one for primary signals and one for conditioning, that converge at the bottleneck via specialized modules and interact through attention mechanisms.

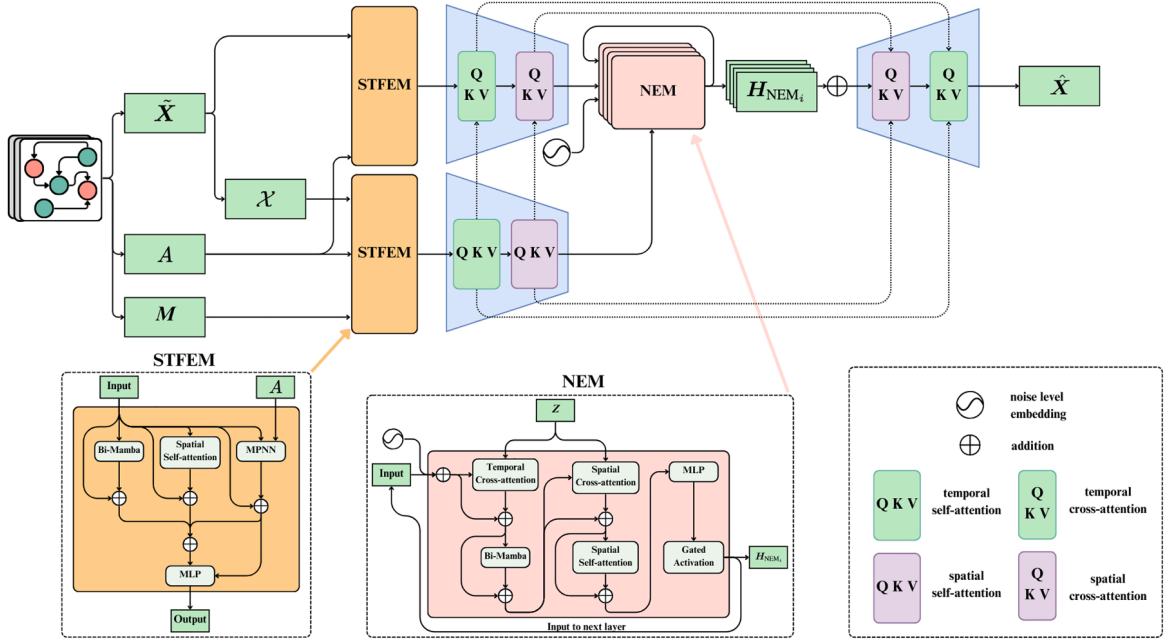

**Fig. 2.** Graphical representation of the architecture implemented in CoSTI. The figure illustrates the two information channels (primary and conditional), the internal structure of the U-Net, and the details of the STFEM and NEM blocks.

Building on this backbone, our hybrid approach integrates the foundational principles of U-Net with recent architectural advances tailored for MTSI [17], enhancing its ability to model complex spatio-temporal dependencies. A high-level overview of the proposed architecture is presented in Fig. 2.

### 4.2.1. Spatio-temporal feature extraction modules (STFEM)

An essential component of our model is the Spatio-Temporal Feature Extraction Module (STFEM), which extends the design principles of the Conditional Feature Extraction Modules proposed in prior work [18]. STFEMs are specifically crafted to extract spatio-temporal representations by combining bidirectional Mamba blocks, transformers with self-attention for spatial dimensions, and a Message Passing Neural Network inspired by Wu et al. [36]. The architecture incorporates two STFEMs corresponding to the model's dual input streams: the primary input channel $(X_t, \mathcal{A})$, located in the upper branch of Fig. 2, and the conditional information channel, positioned in the lower branch, which processes inputs $(\mathcal{X}_t, M_t, \mathcal{A})$.

### 4.2.2. U-Net compression and cross-attention

Following STFEM, both branches of the model proceed through the compression stage of the U-Net architecture. Temporal and spatial dimensions are sequentially reduced by factors $f_t$ and $f_s$, respectively. This follows the "time-then-graph" approach [37], utilizing blocks specifically designed for each dimension and comprising transformers and convolutional layers.

In the conditional branch, information is compressed using self-attention, generating representations at three stages to serve as conditional inputs. In the primary branch, compression employs cross-attention mechanisms to incorporate conditional information. Specifically, we define the primary sequence as $Z \in \mathbb{R}^{n \times d}$, which encodes the original signal path, and the conditional sequence as $H \in \mathbb{R}^{n \times d}$, which comes from the conditioning branch. In each attention block, these are combined using:

$$\text{Attention}(Q, K, V) = \text{softmax}\left(\frac{QK^T}{\sqrt{d}}\right) \cdot V, \quad (8)$$

where $Q = W_Q^{(i)} \cdot Z$, $K = W_K^{(i)} \cdot H$, and $V = W_V^{(i)} \cdot H$, with the matrices $W_Q^{(i)}, W_K^{(i)}, W_V^{(i)} \in \mathbb{R}^{d \times d}$. This mechanism enables the model to condition the processing of the primary sequence on rich contextual information derived from the conditioning branch, aligning both temporal and spatial representations.

### 4.2.3. Noise extraction modules (NEM)

At the bottleneck of the U-Net, Noise Extraction Modules (NEM) are employed, following their introduction by Tashiro et al. [17] and subsequent enhancements by Liu et al. [18] and Solís-García et al. [25]. Originally designed for estimating noise in diffusion models, we adapt NEMs for consistency models, leveraging the conditional information extracted earlier.

These modules integrate bidirectional Mamba blocks in place of temporal transformers. However, given Mamba blocks' lack of native support for cross-attention, we reintroduce transformers for this purpose. Each NEM comprises the following sequence: a temporal transformer with cross-attention for conditional information, a bidirectional Mamba block for sequence analysis, a spatial transformer with cross-attention, and a spatial transformer with self-attention. This design ensures that both temporal and spatial dimensions are analyzed for conditional and intrinsic sequence data.

Finally, the outputs are processed through a Multi-Layer Perceptron (MLP) with gated activation. Each NEM produces two outputs: one as input for the next NEM and another as a noise estimate, $H_{\text{NEM}_i}$. Each block also incorporates an embedding for $\sigma$, constructed using positional embeddings.

### 4.2.4. Final reconstruction

After summing the noise estimates ($H_{\text{NEM}_i}$), the data passes through the reconstruction stage of the U-Net, which restores the original dimensionality of the inputs. This process leverages the conditional information and utilizes skip connections to integrate features from the initial stages of the architecture.

### 4.3. Detailed algorithms description

After presenting the overall architecture, we provide a more detailed description of the algorithms involved in training and sampling, which operationalize the consistency framework described earlier.

The noise scheduler used in this study is based on the Karras scheduler [31], with noise levels $\sigma_i \in \{0.002, 80\}$. The magnitude of $\sigma_i$ and the perturbed input $\tilde{X}_{t,\sigma_i}$ can vary significantly when passed to the model. Therefore, even though it is not always explicitly mentioned in other studies or equations, these inputs are typically scaled for training the neural network responsible for learning the consistency function. Consequently, the inputs are modulated as follows:

$$F_\theta(c_{\text{in}}(\sigma_i) \cdot \tilde{X}_{t,\sigma_i}, \mathcal{X}_t, \mathcal{A}, \tilde{M}_t, c_{\text{noise}}(\sigma_i)) \tag{9}$$

which, following prior work [11,27,31], is implemented in this paper as:

$$c_{\text{in}}(\sigma) = \frac{1}{\sqrt{\sigma^2 + \sigma_{\text{data}}^2}}, \quad c_{\text{in}}(\sigma) = \frac{\ln(\sigma)}{4} \tag{10}$$

To implement the proposed model, we employ two distinct algorithms: one for training (Algorithm 1) and another for sampling or imputation (Algorithm 2).

The training algorithm has a noteworthy feature: it is optimized to skip a forward pass when $\sigma = \sigma_{\min}$. This optimization slightly accelerates execution for small values of $N$, although its impact diminishes as training progresses and $N$ increases.

---

**Algorithm 1** Training algorithm.

**Input:** Dataset $\mathcal{D}$, initial model parameters $\theta$, total steps $K$, currículum scheduler $N(k)$, discrete Lognormal distribution $p(N)$, Pseudo-Huber metric $d(\cdot,\cdot)$ and weighting function $\lambda(\cdot)$
**Init:** $k = 0$
**repeat**
    Sample $(\tilde{X}_t, \tilde{M}_t, M_t, \mathcal{A}_t) \sim \mathcal{D}$, $N \sim N(k)$, $i \sim p(N)$, $\epsilon \sim \mathcal{N}(0,1)$.
    Compute $\tilde{X}_{t,\sigma_i} = \tilde{X}_t + \sigma_i \cdot \epsilon$, $\tilde{X}_{t,\sigma_{i+1}} = \tilde{X}_t + \sigma_{i+1} \cdot \epsilon$
    Compute $\mathcal{X}_t \leftarrow \text{Linear\_interpolation}(\tilde{X}_t)$
    **if** $\sigma_i \neq \sigma_{\min}$ **then**
        $\mathcal{L}(\theta) \leftarrow M_t \cdot \lambda(\sigma_i) d(f_\theta(\tilde{X}_{t,\sigma_{i+1}}, \mathcal{X}_t, \mathcal{A}, \tilde{M}_t, \sigma_{i+1}),$
        $f_{\theta^-}(\tilde{X}_{t,\sigma_i}, \mathcal{X}_t, \mathcal{A}, \tilde{M}_t, \sigma_i))$
                                                        *$f_{\theta^-}$ is done with `stopgrad`*
    **else**
        $\mathcal{L}(\theta) \leftarrow M_t \cdot \lambda(\sigma_i) d(f_\theta(\tilde{X}_{t,\sigma_{i+1}}, \mathcal{X}_t, \mathcal{A}, \tilde{M}_t, \sigma_{i+1}), \tilde{X}_t)$
    **end if**
    $k \leftarrow k + 1$
**until** $k = K$

---

The sampling algorithm supports sampling in as many steps as desired, however in this article we have focused on 1-step and 2-step sampling. By default, sampling begins from the highest noise level allowed, $\sigma_{i_1} = \sigma_{\max} = 80$. For two-step sampling, a dataset-specific noise level $\sigma_{i_2}$ is experimentally determined (details in Section 5.3).

To highlight the efficiency of CoSTI compared to diffusion-based approaches, we analyze the computational complexity of its most demanding components. Given an input tensor of shape $\mathbb{R}^{L \times N \times d}$, where $L$ is the sequence length, $N$ the number of nodes (or variables), and $d$ the feature dimensionality, we focus on the attention modules, which dominate the overall computational cost.

Temporal attention processes each node independently across time, with a per-node cost of $\mathcal{O}(L^2 d)$, resulting in $\mathcal{O}(NL^2 d)$ per layer. Similarly, spatial attention captures interactions across all nodes at each time step, with a cost of $\mathcal{O}(LN^2 d)$ per layer. Other layers, such as feedforward layers or normalization, contribute only marginally and are omitted from this analysis. Combining temporal and spatial attention, the total per-layer complexity is $\mathcal{O}(NL^2 d + LN^2 d)$, and for a model with $B$ blocks, the full forward pass scales as $\mathcal{O}(B(NL^2 d + LN^2 d))$.

Diffusion-based models require $T \gg 1$ iterative denoising steps, each involving a full network evaluation. Their inference complexity is thus $\mathcal{O}(T \cdot B(NL^2 d + LN^2 d))$. In contrast, CoSTI requires only a single forward pass, leading to a theoretical inference speedup of approximately $\sim T$ compared to diffusion-based models. A graphical representation of this concept can be seen in the Fig. 3.

---

**Algorithm 2** Sampling algorithm.

**Input:** Consistency model $f_\theta(\cdot, \cdot)$, sequence of noise levels $\sigma_{i_1} > \sigma_{i_2} > \cdots > \sigma_{i_N}$, data to impute $\tilde{X}_t$, binary missing mask $M_t$, adjacency matrix $\mathcal{A}_t$
**Init:** $S \leftarrow \{\}$         *Set to collect imputations*
Compute $\mathcal{X} \leftarrow \text{Linear\_interpolation}(\tilde{X}_t)$
**for** $j = 1$ **to** 100 **do**
    Sample $\epsilon \sim \mathcal{N}(0,1)$
    $\tilde{X}_{t,\sigma_{i_1}} \leftarrow \tilde{X}_t + \sigma_{i_1} \cdot \epsilon$
    $\hat{X}_t \leftarrow f_\theta(\tilde{X}_{t,\sigma_{i_1}}, \mathcal{X}_t, \mathcal{A}_t, M_t, \sigma_{i_1})$
    **for** $k = 2$ **to** $N$ **do**
        Sample $\epsilon \sim \mathcal{N}(0,1)$
        $\hat{X}_t \leftarrow [\tilde{X}_t \odot M_t + \hat{X}_t \odot (1 - M_t)]$   *keep only missing values*
        $\tilde{X}_{t,\sigma_{i_k}} \leftarrow \hat{X}_t + \sigma_{i_k} \cdot \epsilon$
        $\hat{X}_t \leftarrow f_\theta(\tilde{X}_{t,\sigma_{i_k}}, \mathcal{X}_t, \mathcal{A}_t, \tilde{M}_t, \sigma_{i_k})$
    **end for**
    Add $\hat{X}_t$ to $S$: $S \leftarrow S \cup \{\hat{X}_t\}$
**end for**
$\hat{X}_t \leftarrow \text{Median}(S)$     *Final imputation from median of 100 samples*

---

## 5. Experiments

### 5.1. Datasets

To evaluate our approach, we use datasets that are well-established and frequently cited in the literature, but at the same time serve to represent the importance of obtaining faster imputations.

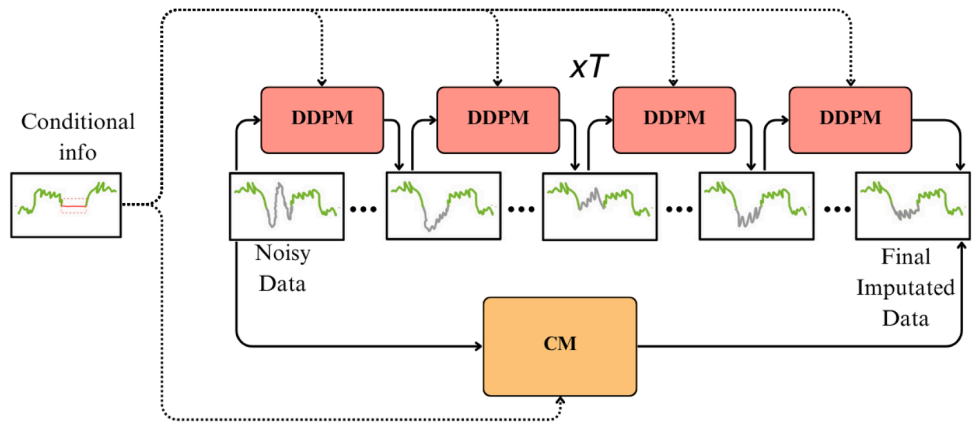

**Fig. 3.** This figure shows the imputation process with a diffusion-based model (DDPM), compared to our consistency-based model (CM). It can be seen how diffusion-based models sample in multiple steps in an iterative way, while consistency models are more efficient as they can sample in a single step.

We used datasets identical to those in Cini et al. [6], which established benchmarks subsequently adopted as comparison metrics. Specifically, the **AQI-36** dataset [38] contains hourly air quality readings from 36 stations in Beijing over one year, with 13.24 % missing data. Additionally, we used the **METR-LA** and **PEMS-BAY** datasets [39], which document traffic conditions in Los Angeles and San Francisco Bay Area, respectively. METR-LA includes records from 207 sensors over four months with an 8.10 % missing rate, while PEMS-BAY comprises data from 325 sensors over six months, with only 0.02 % missing values. These datasets exemplify situations where fast imputation is critical: air quality monitoring ensures timely health alerts, and traffic analysis enables real-time route optimization.

In line with the benchmark, we replicated adjacency matrices using thresholded Gaussian Kernels based on geographical distances, as proposed by Wu et al. [40]. The original data splits (70 % training, 10 % validation, 20 % testing) and random seeds were maintained. For AQI-36, testing occurred in February, May, August, and November, as originally defined, with a 10 % validation sample drawn using the same seed.

We also incorporated data from the **PhysioNet Challenge 2019** [41], which includes clinical records for 40,336 ICU patients from two hospitals. This dataset, aimed at early detection of sepsis, features 40 variables per patient, including vital signs, lab results, and demographic information. Fast imputations are essential in this context, as missing clinical data can delay critical decisions in ICUs. Sepsis requires rapid intervention to reduce mortality, making efficient imputation vital for real-time risk assessments. Following Solís-García et al. [1], we allocated 80 % of the data for training, 10 % for validation, and 10 % for testing. Adjacency matrices were constructed by computing correlations between variables.

In addition, we included the **ETTh1** dataset [42], which consists of hourly temperature and electrical load measurements from an oil-cooled power transformer. The dataset contains 17,420 time steps and 7 features, including oil temperature and electrical parameters. It serves as a representative case of condition monitoring in power systems, where timely imputation of missing sensor values is crucial for predictive maintenance and fault detection. Similarly, we added the **PEMS08** traffic dataset [43], which comprises two months of 5-minute interval traffic readings collected by 170 sensors located in San Bernardino, California. For both ETTh1 and PEMS08, we adopted the same experimental setup as for PhysioNet.

To evaluate imputation performance, we generated synthetic missing data in three scenarios: (1) "Block missing" simulates missing data in 5 % of the samples and masks 1–4-hour blocks at a probability of 0.15 % for each sensor; (2) "Point missing" randomly masks 25 % of values; and (3) a scenario where missing values match the original data distribution.

Table 1 provides detailed information about the datasets, including the topology of their adjacency matrices, the number of nodes and edges, and the proportions of original and additionally injected missing values.

The datasets employed in this study are publicly available and free to use. Specifically, the Torch SpatioTemporal library [44] offers tools to download and preprocess the AQI-36, METR-LA, PEMS-BAY and Pems08 datasets. The Physionet Challenge 2019 dataset is accessible at: https://physionet.org/content/challenge-2019/1.0.0/ and the ETTh-1 is accessible at: https://github.com/zhouhaoyi/ETDataset/blob/main/ETT-small/ETTh1.csv.

### 5.2. Experimental settings

All experiments were conducted five times with different random seeds to ensure robustness. Four synthetic missing data generation strategies were employed during the training:

1) **Point strategy:** Randomly masks [0, 100]% of data per batch. 2) **Block strategy:** Generates sequences of missing values of length [L/2, L] with a [0, 15]% probability, alongside an additional 5 % random missingness. 3) **Historical strategy:** Uses original imputation masks from the dataset. 4) **Hybrid strategy:** Combines the point strategy (primary) with either the block or historical strategy (secondary) with a 50 % probability.

For "Point missing" scenarios, the point strategy was applied. "Block missing" scenarios utilized the hybrid strategy with the block strategy as secondary. For AQI-36, the hybrid strategy was employed with the historical strategy as secondary. Finally, the PhysioNet Challenge 2019, ETTh-1 and Pems08 datasets were trained using the point strategy.

Training epochs varied by dataset: 200 for AQI-36 and ETTh-1; 300 for METR-LA, 150 for PEMS-BAY, and 50 for PhysioNet Challenge 2019 and Pems08. Results for CSDI, PriSTI, and TIMBA were sourced from Solís-García et al. [25], except for the PhysioNet Challenge 2019, where experiments were conducted to obtain comparable results.

Detailed hyperparameters and reproducibility information are provided in Appendix 5.3.

### 5.3. Hyperparameters

This section provides a detailed overview of the hyperparameters used in CoSTI. First, we list the hyperparameters that were kept constant throughout all experiments conducted in this study:

- Batch size: 16
- Learning rate: 2.5e-3
- Channel size $d$: 64
- Number of attention heads: 8
- NEM layers: 4
- $\sigma_{max}$: 80
- $\sigma_{min}$: 0.002
- $\sigma_{data}$: 0.5
- Dropout: 0.2
- $c$: 5.4e-4
- $\rho$: 7
- $s_0$: 10
- $s_1$: 200
- weight decay: 1e-6

**Table 1**
Graph topology and missing value statistics for each dataset.

| Dataset | Graph | | | N. Neighbors | | | % Missing Values | |
|---|---|---|---|---|---|---|---|---|
| | Type | Nodes | Edges | Mean | Median | Isolated Nodes | Original | Additional Injected |
| AQI-36 | undirected | 36 | 343 | 19.06 | 24.5 | 0 | 13.24 | 11.33 |
| METR-LA (P) | directed | 207 | 1515 | 7.32 | 7.0 | 5 | 8.10 | 23.00 |
| (B) | - | - | - | - | - | - | - | 8.4 |
| PEMS-BAY (P) | directed | 325 | 2369 | 7.29 | 7.0 | 12 | 0.02 | 25.0 |
| (B) | - | - | - | - | - | - | - | 9.07 |
| Physionet Challenge 2019 | undirected | 40 | 752 | 37.6 | 38 | 0 | 78.43 | 83.82 |
| ETTh-1 | undirected | 7 | 19 | 5.43 | 5 | 0 | 0 | 25.14 |
| Pems08 | undirected | 170 | 68 | 0.81 | 1.00 | 81 | 0 | 24.99 |

**Table 2**
Hyperparameters of CoSTI for each dataset used in this study.

| Models | AQI-36 | METR-LA | | PEMS-BAY | | Physionet Challenge 2019 | ETTh-1 | Pems08 |
| --- | --- | --- | --- | --- | --- | --- | --- | --- |
| | | Block | Point | Block | Point | | | |
| Time length $L$ | 36 | 24 | 24 | 24 | 24 | 48 | 24 | 24 |
| Epochs | 200 | 300 | 300 | 150 | 150 | 50 | 200 | 50 |
| $f_t$ | 2 | 2 | 1 | 2 | 1 | 2 | 1 | 2 |
| $f_s$ | 2 | 9 | 9 | 5 | 5 | 2 | 1 | 2 |
| $\sigma_{l_2}$ | 20.92 | 0.621 | 0.821 | 1.526 | 5.23 | 20.92 | 0.621 | 0.621 |

**Table 3**
Time in hours required to impute the test set for each dataset using each model. The fastest time is highlighted in bold.

| Dataset | CSDI | PriSTI | TIMBA | CoSTI |
| --- | --- | --- | --- | --- |
| AQI-36 | 0.22 | 0.33 | 0.44 | **0.005** |
| METR-LA | 1.74 | 2.44 | 3.65 | **0.06** |
| PEMS-BAY | 4.62 | 5.99 | 8.71 | **0.16** |
| 1 Physionet | 8.16 | 15.86 | 18.19 | **0.48** |
| ETTh-1 | 0.11 | 0.22 | 0.26 | **0.007** |
| Pems08 | 0.49 | 0.62 | 1.29 | **0.03** |

Table 2 summarizes the hyperparameters that varied across different datasets and evaluation scenarios.

### 5.4. Results and discussion

#### 5.4.1. Performance analysis: Imputation results

Table 3 reports the inference times (in hours) required to impute the test sets of the evaluated datasets for all considered diffusion models and CoSTI using a single sampling step. Additionally, Tables 4 and 5 present the imputation performance of CoSTI with 1 and 2 sampling steps (denoted as CoSTI and CoSTI-2, respectively) compared to other diffusion models across all datasets and experimental scenarios described in Section 5.2.

*Inference speed analysis.* From Table 3, it is evident that CoSTI significantly outperforms diffusion models in terms of inference speed. Unlike CSDI, PriSTI, and TIMBA, which require $T$ steps to generate a sample from noise, CoSTI achieves sampling with just one step. This characteristic makes CoSTI approximately $1/T$ times faster than the other models. For AQI-36, diffusion models use $T = 100$, while for other datasets, $T = 50$, which aligns with the observed results and the complexity analysis from Section 4.3: CoSTI requires only 1.64 % of TIMBA's time for imputing METR-LA and 0.91 % for AQI-36. All experiments were conducted on an NVIDIA RTX A5000 24 GB GPU; further details on hardware and implementation requirements are provided in Appendix A.

*Imputation accuracy analysis.* Tables 4 and 5 highlight that although CoSTI does not consistently achieve the best results, its performance is comparable to that of CSDI, PriSTI, and TIMBA. Notably, CoSTI achieves outstanding results on the Physionet Challenge 2019 dataset. In contrast, CSDI exhibits a significantly higher MSE in this scenario, which can be attributed to convergence failures during training, leading the model to generate predictions close to the mean but with large squared errors. The worst results for Costi are found in the Pems08 dataset. We believe this may be due to a more complex spatio-temporal structure, characterized by many isolated nodes. Therefore, this could potentially be improved with more refined adjacency matrix generation techniques. Interestingly, we observe that increasing the number of sampling steps improves CoSTI's accuracy, as evidenced by the gains obtained with CoSTI-2. This behavior underscores a controllable trade-off between speed and performance, allowing to adapt the model to the specific requirements of the application.

#### 5.4.2. Benchmark analysis

To provide a broader context for the results obtained with CoSTI compared to other imputation techniques, we evaluated it using the comprehensive benchmark introduced by Cini et al. [6]. This benchmark includes a diverse range of methods, encompassing both traditional statistical approaches and more advanced generative models. Specifically, it evaluates the following categories of techniques:

1. **Statistical techniques** Methods based on simple aggregations, such as MEAN (historical average values), DA (daily averages), and KNN (proximity-based imputation).
2. **Machine learning algorithms** These include Lin-ITP (linear interpolation), MICE [45] (multiple imputations), VAR (vector autoregressive models), and KF (Kalman Filter).
3. **Matrix factorization techniques** TRMF [46] (temporal regularized matrix factorization) and BATF [47] (Bayesian augmented tensor factorization).
4. **Autoregressive models** These include BRITS [5], MPGRU (a GNN-based predictor resembling DCRNN [39]), and GRIN [6].
5. **Generative models** Advanced generative techniques, such as CSDI, PriSTI, TIMBA, V-RIN [48] (a VAE with uncertainty quantification), GP-VAE [20] (a combination of VAEs and Gaussian processes), and rGAIN [21] (GAN-based imputation with recurrent structures).

The detailed evaluation results, summarized in Table 6, highlight the strong performance of diffusion-based approaches such as CSDI, PriSTI, and TIMBA. Notably, CoSTI achieves comparable or superior results in multiple cases, demonstrating its ability to harness the strengths of

**Table 4**
Results obtained by applying the models to the AQI-36, PhysioNet Challenge 2019, ETTh-1 and Pems08 datasets with additional synthetic missing values. The best value is highlighted in bold, and any case where CoSTI outperforms a DDPM model is underlined.

| | AQI-36 | | Physionet Challenge 2019 | | ETTh-1 | | Pems08 | |
| --- | --- | --- | --- | --- | --- | --- | --- | --- |
| | Simulated failure (24.6 %) | | Simulated failure (83.82 %) | | Simulated failure (25.14 %) | | Simulated failure (24.99 %) | |
| Models | MAE | MSE | MAE | MSE | MAE | MSE | MAE | MSE |
| CSDI | 9.74 ± 0.16 | 388.37 ± 11.42 | 3.93 ± 0.76 | 2282.94 ± 2646.26 | **0.33** ± 0.00 | 0.42 ± 0.01 | 9.87 ± 0.01 | 293.31 ± 1.10 |
| PriSTI | 9.84 ± 0.11 | 376.11 ± 10.62 | 3.58 ± 0.33 | 573.06 ± 22.89 | **0.33** ± 0.00 | 0.41 ± 0.01 | 9.93 ± 0.05 | **289.28** ± 3.38 |
| TIMBA | **9.56** ± 0.4 | **352.29** ± 5.33 | 3.11 ± 0.32 | 521.29 ± 23.11 | **0.33** ± 0.00 | 0.41 ± 0.01 | 9.97 ± 0.04 | 293.95 ± 2.49 |
| CoSTI | 10.13 ± 0.08 | 377.48 ± 9.29 | 2.47 ± 0.05 | 388.31 ± 36.55 | 0.38 ± 0.00 | 0.48 ± 0.01 | 11.09 ± 0.27 | 336.93 ± 10.21 |
| CoSTI-2 | 9.90 ± 0.13 | 358.67 ± 13.05 | **2.46** ± 0.07 | **334.40** ± 23.84 | 0.35 ± 0.01 | 0.43 ± 0.02 | 10.78 ± 0.04 | 329.14 ± 2.89 |

**Table 5**

Results obtained by applying the models to the METR-LA and PEMS-BAY datasets under the "Point missing" and "Block missing" scenarios. The best value is highlighted in bold, and any case where CoSTI outperforms a DDPM model is underlined.

| | METR-LA | | | | PEMS-BAY | | | |
|---|---|---|---|---|---|---|---|---|
| | Block-missing (16.6%) | | Point-missing (31.1%) | | Block-missing (9.2%) | | Point-missing (25.0%) | |
| Models | MAE | MSE | MAE | MSE | MAE | MSE | MAE | MSE |
| CSDI | 1.90 ± 0.01 | 12.27 ± 0.18 | 1.77 ± 0.05 | 9.42 ± 0.47 | **0.84** ± 0.00 | **4.06** ± 0.04 | **0.58** ± 0.00 | **1.30** ± 0.04 |
| PriSTI | 1.78 ± 0.00 | 10.64 ± 0.13 | 1.70 ± 0.00 | 8.47 ± 0.04 | 0.87 ± 0.01 | 4.64 ± 0.21 | 0.59 ± 0.00 | 1.61 ± 0.03 |
| TIMBA | **1.76** ± 0.02 | **10.36** ± 0.34 | **1.69** ± 0.00 | **8.36** ± 0.01 | **0.84** ± 0.01 | 4.57 ± 0.08 | **0.58** ± 0.00 | 1.63 ± 0.08 |
| CoSTI | 1.85 ± 0.01 | 11.44 ± 0.08 | 1.76 ± 0.01 | 9.01 ± 0.06 | 0.93 ± 0.03 | 4.52 ± 0.15 | 0.64 ± 0.01 | 1.53 ± 0.15 |
| CoSTI-2 | 1.84 ± 0.01 | 11.35 ± 0.14 | 1.75 ± 0.00 | 8.97 ± 0.05 | 0.91 ± 0.02 | 4.43 ± 0.12 | 0.63 ± 0.01 | 1.44 ± 0.12 |

**Table 6**

Performance comparison of CoSTI with existing methods from the benchmark established in the literature [6]. Results are presented in terms of MAE and MSE, providing a comprehensive evaluation of imputation accuracy. The best value is highlighted in bold, and any case where CoSTI outperforms a DDPM model is underlined.

| | AQI-36 | | METR-LA | | | | PEMS-BAY | | | |
|---|---|---|---|---|---|---|---|---|---|---|
| | Simulated failure (24.6%) | | Block-missing (16.6%) | | Point-missing (31.1%) | | Block-missing (9.2%) | | Point-missing (25.0%) | |
| Models | MAE | MSE | MAE | MSE | MAE | MSE | MAE | MSE | MAE | MSE |
| Mean | 53.48 ± 0.00 | 4578.08 ± 0.00 | 7.48 ± 0.00 | 139.54 ± 0.00 | 7.56 ± 0.00 | 142.22 ± 0.00 | 5.46 ± 0.00 | 87.56 ± 0.00 | 5.42 ± 0.00 | 86.59 ± 0.00 |
| DA | 50.51 ± 0.00 | 4416.10 ± 0.00 | 14.53 ± 0.00 | 445.08 ± 0.00 | 14.57 ± 0.00 | 448.66 ± 0.00 | 3.30 ± 0.00 | 43.76 ± 0.00 | 3.35 ± 0.00 | 44.50 ± 0.00 |
| KNN | 30.21 ± 0.00 | 2892.31 ± 0.00 | 7.79 ± 0.00 | 124.61 ± 0.00 | 7.88 ± 0.00 | 129.29 ± 0.00 | 4.30 ± 0.00 | 49.90 ± 0.00 | 4.30 ± 0.00 | 49.80 ± 0.00 |
| Lin-ITP | 14.46 ± 0.00 | 673.92 ± 0.00 | 3.26 ± 0.00 | 33.76 ± 0.00 | 2.43 ± 0.00 | 14.75 ± 0.00 | 1.54 ± 0.00 | 14.14 ± 0.00 | 0.76 ± 0.00 | 1.74 ± 0.00 |
| KF | 54.09 ± 0.00 | 4942.26 ± 0.00 | 16.75 ± 0.00 | 534.69 ± 0.00 | 16.66 ± 0.00 | 529.96 ± 0.00 | 5.64 ± 0.00 | 93.19 ± 0.00 | 5.68 ± 0.00 | 93.32 ± 0.00 |
| MICE | 30.37 ± 0.09 | 2594.06 ± 7.17 | 4.22 ± 0.05 | 51.07 ± 1.25 | 4.42 ± 0.07 | 55.07 ± 1.46 | 2.94 ± 0.02 | 28.28 ± 0.37 | 3.09 ± 0.02 | 31.43 ± 0.41 |
| VAR | 15.64 ± 0.08 | 833.46 ± 13.85 | 3.11 ± 0.08 | 28.00 ± 0.76 | 2.69 ± 0.00 | 21.10 ± 0.02 | 2.09 ± 0.10 | 16.06 ± 0.73 | 1.30 ± 0.00 | 6.52 ± 0.01 |
| TRMF | 15.46 ± 0.06 | 1379.05 ± 34.83 | 2.96 ± 0.00 | 22.65 ± 0.13 | 2.86 ± 0.00 | 20.39 ± 0.02 | 1.95 ± 0.01 | 11.21 ± 0.06 | 1.85 ± 0.00 | 10.03 ± 0.00 |
| BATF | 15.21 ± 0.27 | 662.87 ± 29.55 | 3.56 ± 0.01 | 35.39 ± 0.03 | 3.58 ± 0.01 | 36.05 ± 0.02 | 2.05 ± 0.00 | 14.48 ± 0.01 | 2.05 ± 0.00 | 14.90 ± 0.06 |
| V-RIN | 10.00 ± 0.10 | 838.05 ± 24.74 | 6.84 ± 0.17 | 150.08 ± 6.13 | 3.96 ± 0.08 | 49.98 ± 1.30 | 2.49 ± 0.04 | 36.12 ± 0.66 | 1.21 ± 0.03 | 6.08 ± 0.29 |
| GP-VAE | 25.71 ± 0.30 | 2589.53 ± 59.14 | 6.55 ± 0.09 | 122.33 ± 2.05 | 6.57 ± 0.10 | 127.26 ± 3.97 | 2.86 ± 0.15 | 26.80 ± 2.10 | 3.41 ± 0.23 | 38.95 ± 4.16 |
| rGAIN | 15.37 ± 0.26 | 641.92 ± 33.89 | 2.90 ± 0.01 | 21.67 ± 0.15 | 2.83 ± 0.01 | 20.03 ± 0.09 | 2.18 ± 0.01 | 13.96 ± 0.20 | 1.88 ± 0.02 | 10.37 ± 0.20 |
| MPGRU | 16.79 ± 0.52 | 1103.04 ± 106.83 | 2.57 ± 0.01 | 25.15 ± 0.17 | 2.44 ± 0.00 | 22.17 ± 0.03 | 1.59 ± 0.01 | 14.19 ± 0.11 | 1.11 ± 0.00 | 7.59 ± 0.02 |
| BRITS | 14.50 ± 0.35 | 622.36 ± 65.16 | 2.34 ± 0.01 | 17.00 ± 0.14 | 2.34 ± 0.00 | 16.46 ± 0.05 | 1.70 ± 0.01 | 10.50 ± 0.07 | 1.47 ± 0.00 | 7.94 ± 0.03 |
| GRIN | 12.08 ± 0.47 | 523.14 ± 57.17 | 2.03 ± 0.00 | 13.26 ± 0.05 | 1.91 ± 0.00 | 10.41 ± 0.03 | 1.14 ± 0.01 | 6.60 ± 0.10 | 0.67 ± 0.00 | 1.55 ± 0.01 |
| CSDI | 9.74 ± 0.16 | 388.37 ± 11.42 | 1.90 ± 0.01 | 12.27 ± 0.18 | 1.77 ± 0.05 | 9.42 ± 0.47 | **0.84** ± 0.00 | **4.06** ± 0.04 | **0.58** ± 0.00 | **1.30** ± 0.04 |
| PriSTI | 9.84 ± 0.11 | 376.11 ± 10.62 | 1.78 ± 0.00 | 10.64 ± 0.13 | 1.70 ± 0.00 | 8.47 ± 0.04 | 0.87 ± 0.01 | 4.64 ± 0.21 | 0.59 ± 0.00 | 1.61 ± 0.03 |
| TIMBA | **9.56** ± 0.4 | **352.29** ± 5.33 | **1.76** ± 0.02 | **10.36** ± 0.34 | **1.69** ± 0.00 | **8.36** ± 0.01 | **0.84** ± 0.01 | 4.57 ± 0.08 | **0.58** ± 0.00 | 1.63 ± 0.08 |
| CoSTI | 10.13 ± 0.08 | 377.48 ± 9.29 | 1.85 ± 0.01 | 11.44 ± 0.08 | 1.76 ± 0.01 | 9.01 ± 0.06 | 0.93 ± 0.03 | 4.52 ± 0.15 | 0.64 ± 0.01 | 1.53 ± 0.15 |
| CoSTI-2 | 9.90 ± 0.13 | 358.67 ± 13.05 | 1.84 ± 0.01 | 11.35 ± 0.14 | 1.75 ± 0.00 | 8.97 ± 0.05 | 0.91 ± 0.02 | 4.43 ± 0.12 | 0.63 ± 0.01 | 1.44 ± 0.12 |

diffusion models, particularly their capacity to outperform traditional methods like autoregressive techniques [17].

These findings emphasize the competitiveness of CoSTI across a diverse benchmark of imputation techniques. While diffusion-based models, including CSDI, PriSTI, and TIMBA, consistently deliver state-of-the-art performance, CoSTI matches their MSE across various datasets and imputation scenarios. However, in certain cases, CoSTI exhibits slightly higher MAE compared to some diffusion-based counterparts, suggesting potential areas for improvement in capturing absolute error dynamics. Despite this, CoSTI demonstrates robust and versatile performance, positioning itself as a strong alternative to leading diffusion-based models, particularly in scenarios where balancing accuracy and speed is crucial. To better understand this balance, we examine the trade-off between computational efficiency and imputation accuracy across all evaluated methods.

Fig. 4 reveals the fundamental trade-off between imputation accuracy and computational efficiency, providing deeper insight into the competitive positioning established in our benchmark results. Traditional approaches present a clear dichotomy: diffusion-based models achieve the lowest MAE values through iterative sampling but at substantial computational cost, while non-diffusion models offer fast inference but with consistently limited accuracy.

CoSTI breaks this traditional compromise by occupying a previously unexplored region of the accuracy-speed space. While not always matching diffusion models' absolute performance, CoSTI consistently outperforms all non-diffusion approaches while maintaining computational efficiency comparable to fast methods. The two-step variant, CoSTI-2, extends this advantage further, achieving better accuracy with only modest increases in inference time.

Critically, both CoSTI variants consistently appear on the Pareto frontier across all benchmarks, the optimal boundary where no solution can improve accuracy without sacrificing speed, or vice versa. This positioning represents a paradigm shift: CoSTI delivers the quality previously exclusive to expensive diffusion models while preserving the practical inference speeds essential for real-world deployment.

*5.4.3. Sensitivity analysis*

To evaluate sensitivity to varying missing value ratios, we tested CoSTI, CSDI, PriSTI, and TIMBA on the METR-LA dataset under the Point-missing scenario with different levels of missing data. Leveraging the best-performing model weights from Table 5, the results are illustrated in Fig. 5, with further details provided in Tables 7 and 8. CoSTI exhibits performance comparable to PriSTI and TIMBA, particularly under higher missing rates, underscoring its robustness and reliability in challenging imputation scenarios.

*5.4.4. Downstream task evaluation*

Assessing imputation models solely on reconstruction metrics may not reflect their practical utility. Thus, we evaluate their effectiveness on a downstream task: node value prediction. This task, consistent with Solís-García et al. [25], involves predicting a node's value at time *t* using values from other nodes at the same timestamp.

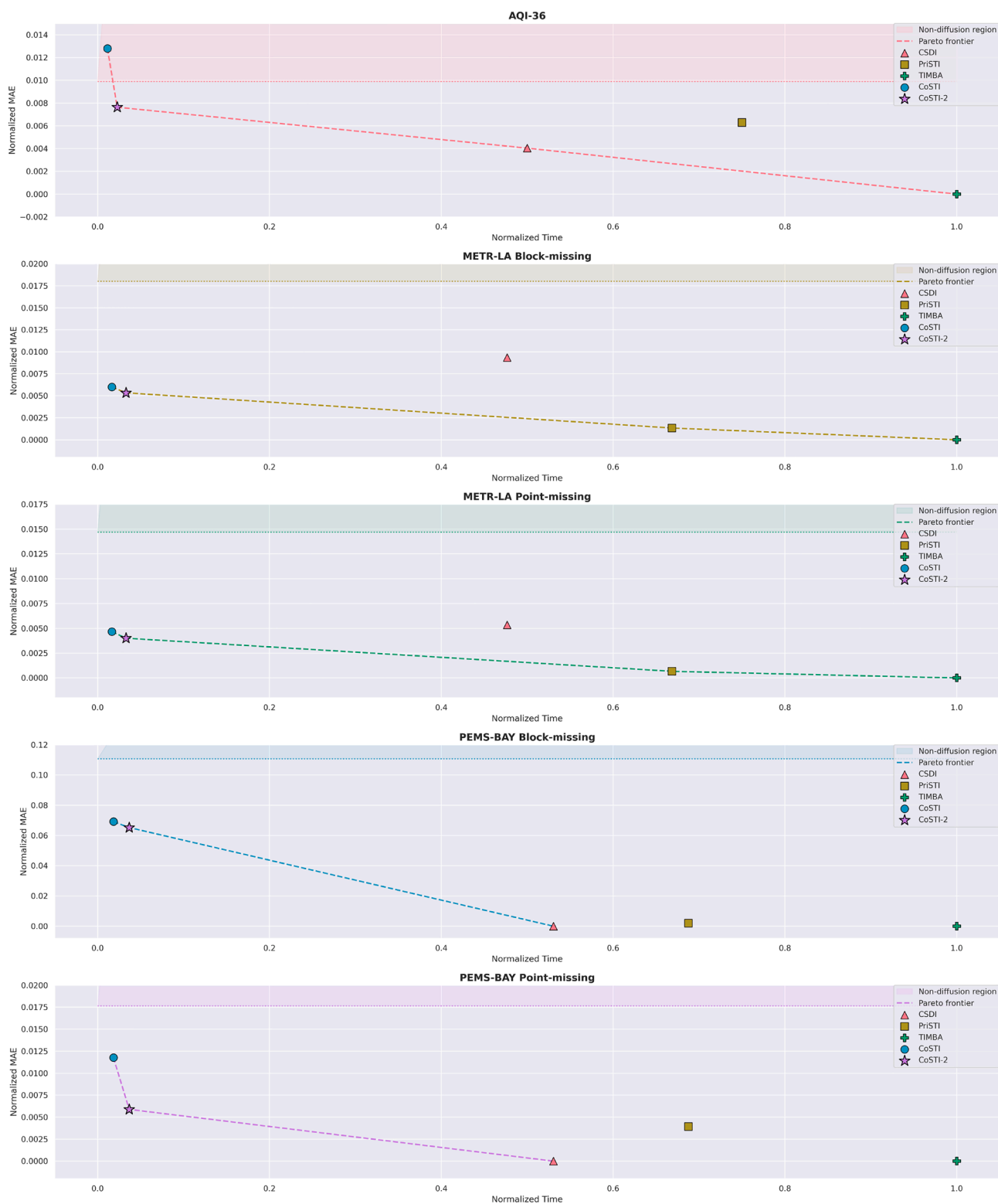

**Fig. 4.** Pareto frontier of normalized MAE versus normalized inference time across the imputation scenarios of the benchmark. Each MAE value is normalized using the minimum and maximum MAE from Table 6 for the corresponding dataset. Similarly, inference times are normalized using the minimum and maximum reported values for each dataset, as summarized in Table 3. Dashed lines indicate the Pareto frontiers, and shaded regions highlight the performance band of non-diffusion models.

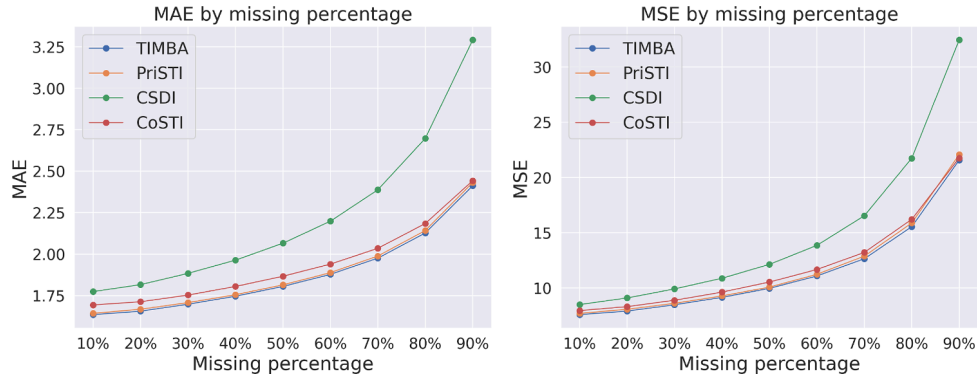

**Fig. 5.** Graphs showing the performance of the models under different missing value ratios. Since PriSTI and TIMBA overlap significantly, making them harder to distinguish.

**Table 7**
Sensitivity analysis for different levels of missing values in the METR-LA dataset under the Point missing scenario, presented in terms of MAE. The best value is highlighted in bold, and any case where CoSTI outperforms a DDPM model is underlined.

| | MAE - METR-LA (P) | | | | | | | | |
|---|---|---|---|---|---|---|---|---|---|
| Models | 10% | 20% | 30% | 40% | 50% | 60% | 70% | 80% | 90% |
| CSDI | 1.77 | 1.82 | 1.88 | 1.96 | 2.07 | 2.20 | 2.39 | 2.70 | 3.29 |
| PriSTI | **1.64** | 1.67 | 1.71 | 1.76 | 1.81 | 1.89 | 1.99 | 2.14 | 2.43 |
| TIMBA | **1.64** | **1.66** | **1.70** | **1.75** | **1.80** | **1.88** | **1.98** | **2.13** | **2.41** |
| CoSTI | <u>1.69</u> | <u>1.71</u> | <u>1.75</u> | <u>1.80</u> | <u>1.87</u> | <u>1.94</u> | <u>2.03</u> | <u>2.18</u> | <u>2.44</u> |

**Table 8**
Sensitivity analysis for different levels of missing values in the METR-LA dataset under the Point missing scenario, presented in terms of MSE. The best value is highlighted in bold, and any case where CoSTI outperforms a DDPM model is underlined.

| | MSE - METR-LA (P) | | | | | | | | |
|---|---|---|---|---|---|---|---|---|---|
| Models | 10% | 20% | 30% | 40% | 50% | 60% | 70% | 80% | 90% |
| CSDI | 8.50 | 9.10 | 9.92 | 10.88 | 12.14 | 13.86 | 16.54 | 21.75 | 32.54 |
| PriSTI | 7.71 | 8.07 | 8.61 | 9.29 | 10.08 | 11.24 | 12.88 | 15.88 | 22.08 |
| TIMBA | **7.60** | **7.91** | **8.48** | **9.15** | **9.96** | **11.09** | **12.65** | **15.55** | **21.57** |
| CoSTI | <u>7.95</u> | <u>8.31</u> | <u>8.89</u> | <u>9.63</u> | <u>10.54</u> | <u>11.67</u> | <u>13.22</u> | <u>16.22</u> | <u>21.77</u> |

Using the AQI-36 dataset, missing values were imputed with the best-performing model weights from Table 4. Imputed data were split into a new training (80 %) and test set (20 %), normalized with a Min-Max scaler. We trained an MLP with one hidden layer (100 neurons) for 500 epochs, minimizing MSE. Final test metrics excluded imputed target values, focusing on actual values only. As seen in Table 9, CoSTI achieves comparable performance to diffusion-based models, outperforming them in node 31 prediction.

**Table 9**
Performance results for the downstream task of node value prediction using data imputed by the evaluated models. The best value is highlighted in bold, and any case where CoSTI outperforms a DDPM model is underlined.

| | Sensor 14 | | Sensor 31 | |
|---|---|---|---|---|
| Models | MAE | MSE | MAE | MSE |
| CSDI | 6.51 ± 0.69 | 96.99 ± 21.25 | 11.99 ± 1.92 | 376.40 ± 148.06 |
| PriSTI | 6.46 ± 0.71 | 92.70 ± 20.27 | 11.70 ± 1.80 | 361.19 ± 132.99 |
| TIMBA | **6.45** ± 0.69 | **91.90** ± 20.33 | 11.68 ± 1.77 | 359.80 ± 131.74 |
| CoSTI | <u>6.51</u> ± 0.67 | <u>93.96</u> ± 20.60 | **<u>11.41</u>** ± 1.88 | **<u>350.46</u>** ± 135.60 |

*5.4.5. Architectural component ablations*
We conducted the following experiments to analyze the impact of architectural components:

- **w/o Cond**: This experiment evaluates the performance of CoSTI when the second model head, responsible for extracting conditional information, is removed.
- **w/o SFTE**: This test assesses the model's performance when the STFEM modules, which extract spatio-temporal representations of the input sequence at the network's entry, are omitted. D
- **w/o NEM**: This experiment examines the impact of removing the NEM blocks, located at the core of the network, that are designed to estimate noise in the input sequence.
- **w/o self-attention**: This test evaluates the contribution of the Bi-Mamba and Spatial Self-Attention layers within the NEM blocks to assess whether their inclusion enhances performance.

The results in Table 10 demonstrate the contributions of the components evaluated through the ablation study. The removal of the conditional head (**w/o Cond**) results in the largest performance degradation, with substantial increases in both MAE and MSE. This highlights the importance of the conditional information captured by this head, which appears crucial for the model's ability to make accurate predictions. Similarly, excluding the STFEM module (**w/o SFTE**) considerably worsens performance, indicating that the spatio-temporal representations extracted by this module are fundamental for capturing the input sequence's underlying structure.

In the case of the NEM blocks (**w/o NEM**), their removal causes a slight performance decline, which suggests that noise estimation primarily contributes to refining the model's predictions but is not as critical as other components. Likewise, excluding the self-attention mechanism (**w/o self-attention**) leads to marginally higher MAE and MSE. Although the difference is small, this indicates that the Bi-Mamba and Spatial Self-Attention layers enhance the model's ability to focus on relevant spatio-temporal features, leading to slightly better performance. These results collectively emphasize that each component, even those with smaller individual contributions, plays a role in improving the model's overall robustness and precision.

**Table 10**
Performance comparison (MAE and MSE) of CoSTI and its ablated versions across datasets.

| Dataset | MAE | MSE |
|---|---|---|
| CoSTI | **1.76** ± 0.01 | **9.01** ± 0.06 |
| w/o Cond | 6.36 ± 0.02 | 153.51 ± 1.89 |
| w/o STFE | 2.33 ± 0.06 | 16.32 ± 0.99 |
| w/o NEM | 1.77 ± 0.01 | 9.17 ± 0.05 |
| w/o self-attention | 1.77 ± 0.01 | 9.14 ± 0.14 |

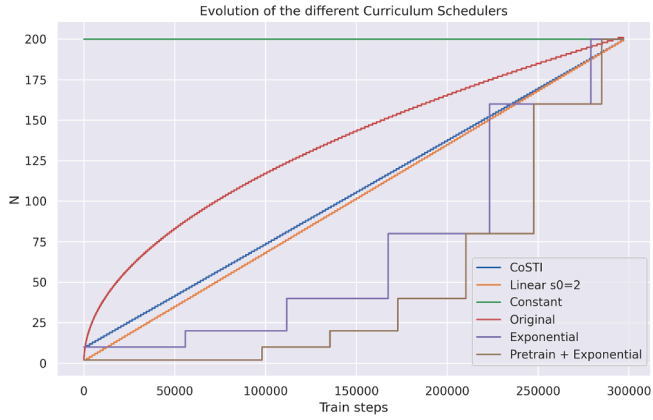

**Fig. 6.** Graphical representation of selected Curriculum Schedulers evaluated in this experiment.

*5.4.6. Curriculum learning strategy evaluation*

In this section, we explore various curriculum learning schedulers to evaluate their impact on model performance, with a graphical representation of some of the tested schedulers provided in Fig. 6.

- **CoSTI scheduler**: This is the scheduler implemented in this paper. It is a linear scheduler with $s_0 = 10$ and $s_1 = 200$. In Table 11 and Fig. 6, it is labeled as "Linear $s_0 = 10$."
- **Linear $s_0 = 2$**: This scheduler is similar to the one implemented in this paper but uses $s_0 = 2$.
- **Linear $s_1 = 1280$**: This scheduler also follows a linear approach, but with $s_0 = 10$ and $s_1 = 1280$, as proposed in Song and Dhariwal [26].
- **Original scheduler**: This scheduler is the one originally proposed in Song et al. [11], with $s_0 = 2$ and $s_1 = 200$.
- **Constant**: This is the simplest scheduler, where $N = 200$ remains constant throughout training.
- **Exponential**: Proposed in Song and Dhariwal [26], this scheduler increases $N$ exponentially until reaching the target value, with $s_0 = 10$ and $s_1 = 200$.
- **Pretrain + Exponential**: This is a custom scheduler that introduces a pretraining phase with $N = 2$. Inspired by Geng et al. [27], this phase aims to train the model to generate accurate imputations for MTS starting from pure noise by comparing the noisy input with the original unperturbed sample. This process helps the model learn to predict correct imputations before being trained for consistency across different $\sigma$ values. After this pretraining phase, it is an exponential scheduler with $s_0 = 10$ and $s_1 = 200$. The pretraining phase comprises $r = 1/3$ of the total steps, followed by the exponential scheduler.

The results in Table 11 highlight why the linear scheduler with $s_0 = 10$ and $s_1 = 200$ achieves such competitive performance, even when compared to more elaborate approaches like exponential growth or pretraining-based strategies. This success stems from the same principles that motivate these advanced techniques: a progressive improvement in model performance as $N$ increases. While larger $N$ values reduce approximation bias and improve imputation quality, they can also introduce higher variance or instability during training [26,27]. The linear scheduler effectively balances this trade-off by providing a gradual, steady increase in $N$. Unlike exponential approaches, which can transition too aggressively, the linear strategy ensures that each $N$ range is explored sufficiently, allowing the model to adapt to incremental increases in difficulty without destabilizing the training process.

Furthermore, the simplicity of the linear scheduler is a significant advantage. More complex strategies, such as pretraining followed by exponential growth, introduce additional design considerations, such as determining pretraining duration or tuning exponential rates. By contrast, the linear scheduler only requires setting an initial and final value for $N$. This simplicity reduces the risk of implementation errors and minimizes the burden of hyperparameter tuning, making it a practical choice for real-world applications. Additionally, starting at $N = 10$ offers a moderate initial difficulty level that allows the model to learn foundational patterns effectively before progressing to more complex tasks. We theorize that this moderate starting point strikes a good balance between simplicity and effectiveness, ensuring a smooth progression during training.

**Table 11**
Performance results for different curriculum learning schedulers.

| Dataset | MAE | MSE |
|---|---|---|
| Linear $s_0 = 10$ (CoSTI) | **1.76** ± 0.01 | **9.01** ± 0.06 |
| Linear $s_0 = 2$ | 1.87 ± 0.05 | 10.01 ± 0.43 |
| Linear $s_1 = 1280$ | 1.82 ± 0.01 | 9.46 ± 0.19 |
| Constant | 1.86 ± 0.03 | 9.30 ± 0.29 |
| Original | 1.80 ± 0.02 | 9.30 ± 0.28 |
| Exponential | 1.79 ± 0.03 | 9.44 ± 0.41 |
| Pretrain + Exponential | 1.79 ± 0.03 | 9.18 ± 0.07 |

**Table 12**
Performance results for different optimizers.

| Dataset | MAE | MSE |
|---|---|---|
| AdamWScheduleFree (CoSTI) | **1.76** ± 0.01 | **9.01** ± 0.06 |
| RAdam | 1.84 ± 0.02 | 9.99 ± 0.31 |
| AdamW | 1.81 ± 0.02 | 9.41 ± 0.14 |

While exponential and pretraining strategies have strong theoretical justifications, the linear scheduler with $s_0 = 10$ and $s_1 = 200$ demonstrates that simplicity and stability can often yield equally strong, if not superior, results. Its gradual and controlled progression strikes an effective balance between bias and variance, ensuring robust model adaptation throughout training. This combination of simplicity and progressive adaptation makes the linear approach not only a competitive alternative but also a reliable and efficient solution for MTSI tasks.

*5.4.7. Optimizer performance comparison*

In this section, we compare the original optimizer used in our experiments, **AdamWScheduleFree** [34], with two alternative configurations widely used in the literature:

- **RAdam**: This optimizer is commonly employed in papers on Consistency Models [11,26,27]. The configuration uses the RAdam optimizer [49] without learning rate decay, warm-up, or weight decay. The learning rate is set to 1e-4, as suggested in Song and Dhariwal [26].
- **AdamW + MultiStepLR**: This configuration is used in models such as CSDI, PriSTI, and TIMBA. It combines AdamW with a learning rate of 1e-3 and weight decay of 1e-6. Additionally, it employs a MultiStepLR schedule, reducing the learning rate to 1e-4 and 1e-5 when 75 % and 90 % of training epochs are completed, respectively.

The comparison of optimizers highlights the efficacy of the proposed AdamWScheduleFree configuration in achieving superior quantitative performance and maintaining training stability. As shown in Table 12, AdamWScheduleFree consistently outperforms the alternatives, achieving the lowest MAE and MSE across experiments. While AdamW achieves slightly better results than RAdam, its reliance on a complex learning rate schedule introduces additional hyperparameter tuning overhead, which is avoided by the simpler, schedule-free design of AdamWScheduleFree.

Fig. 7 provides further insights into the training dynamics. AdamWScheduleFree achieves smoother and more consistent convergence in both training and validation loss curves, maintaining a lower imputation error on the validation set (measured in MAE) throughout the training process. The progressive smoothing of the training loss can be attributed to the curriculum learning strategy employed. By incrementally increasing the number of $N$, the PF-FLOW comparisons are progressively localized to closer points, which reduces the complexity of the task over time. This design allows the model to gradually refine its predictions, leading to a steady and significant reduction in training loss.

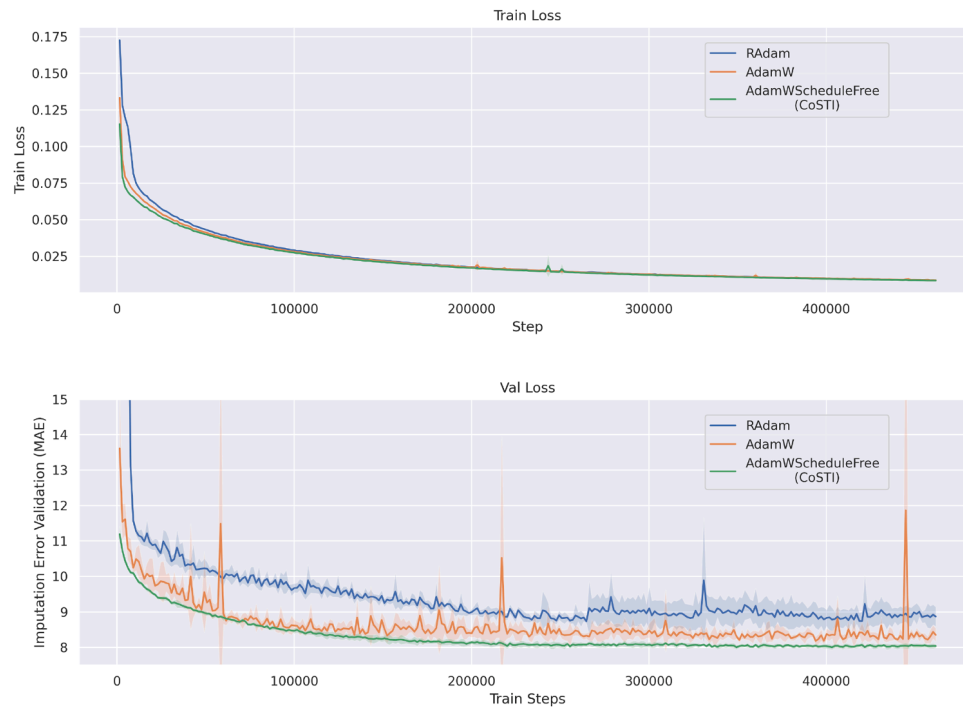

**Fig. 7.** Training and validation loss curves for the optimizer comparison experiment. Shaded regions indicate the mean and standard deviation across five runs, illustrating the consistency and variability of each optimizer's performance.

Overall, these results demonstrate that AdamWScheduleFree simplifies the optimization process while delivering state-of-the-art performance. Its ability to achieve competitive results highlights its effectiveness and practicality for imputation tasks, making it a strong alternative to more complex configurations.

## 6. Concusions

This work introduced CoSTI, a novel model for MTSI that leverages Consistency Models to match the performance of state-of-the-art diffusion-based methods while drastically reducing inference time. Through the use of Consistency Training and domain-specific conditional signals, CoSTI achieves up to a 98 % reduction in imputation time across diverse datasets and missingness scenarios.

These results position CoSTI as an efficient and scalable solution for real-time applications in domains such as healthcare and traffic monitoring, where fast and accurate imputations are essential. To the best of our knowledge, this is the first adaptation of Consistency Models to the MTSI setting, opening new directions for fast generative modeling in time series domains.

While the model shows promising results, several aspects remain open for future exploration. First, although this study focuses on imputation, the underlying formulation of CoSTI is naturally extendable to related tasks such as time series forecasting or anomaly detection. Second, our approach adopts a fixed graph structure to represent spatial dependencies, an effective and widely used design choice, but one that could be enhanced by incorporating dynamic or learnable graphs in scenarios with evolving topologies. Third, we observed that training stability can vary depending on the dataset and initialization, suggesting opportunities to improve robustness through better curriculum strategies or initialization schemes.

Future work will also consider optimizing the noise schedule and number of sampling steps per dataset, and investigating latent consistency models to further accelerate training. Overall, CoSTI provides a strong foundation for scalable and fast time series modeling and opens the door to a broader class of generative solutions in structured temporal domains.

## Declaration of generative AI and AI-assisted technologies in the writing process

During the preparation of this work the author(s) used ChatGPT in order to improve writing and check for language errors. After using this tool/service, the author(s) reviewed and edited the content as needed and take(s) full responsibility for the content of the publication.

## CRediT authorship contribution statement

**Javier Solís-García:** Writing – review & editing, Writing – original draft, Visualization, Software, Project administration, Methodology, Investigation, Formal analysis, Data curation, Conceptualization; **Belén Vega-Márquez:** Writing – review & editing, Supervision; **Juan A. Nepomuceno:** Writing – review & editing, Formal analysis; **Isabel A. Nepomuceno-Chamorro:** Writing – review & editing, Supervision, Funding acquisition, Formal analysis.

## Data availability

The link to data and code is provided in the manuscript

## Declaration of competing interest

The authors declare that they have no known competing financial interests or personal relationships that could have appeared to influence the work reported in this paper.

## Acknowledgements

This work has been funding by the Spanish Ministry of Science and Innovation under project PID2020-117954RB-C22 financed by MCIN/ AEI /10.13039/501100011033.

**Table A.13**
Training and inference times measured in hours.

|  | AQI-36 | | METR-LA | | PEMS-BAY | | Physionet Challenge 2019 | | EETh-1 | | Pems08 | |
| --- | --- | --- | --- | --- | --- | --- | --- | --- | --- | --- | --- | --- |
| Models | Training | Inference | Training | Inference | Training | Inference | Training | Inference | Training | Inference | Training | Inference |
| CSDI | 1.17 | 0.22 | 43.67 | 1.74 | 114.67 | 4.62 | 7.35 | 8.16 | 0.18 | 0.11 | 2.68 | 0.49 |
| PriSTI | 1.61 | 0.33 | 49.25 | 2.44 | 118.33 | 5.99 | 13.13 | 15.87 | 0.31 | 0.22 | 3.19 | 0.62 |
| TIMBA | 0.56 | 0.44 | 77.00 | 3.65 | 196.42 | 8.71 | 15.03 | 18.19 | 0.37 | 0.26 | 5.04 | 1.29 |
| CoSTI | 1.99 | 0.005 | 79.42 | 0.06 | 108.75 | 0.16 | 12.92 | 0.48 | 0.33 | 0.007 | 4.50 | 0.03 |

## Appendix A. Implementation details

### A.1. Code reproducibility

This study emphasizes code reproducibility and accessibility to the datasets employed. The implementation has been made publicly available on GitHub: https://github.com/javiersgjavi/CoSTI.

The codebase is written in Python [50] and leverages several widely used open-source libraries, including:

- Pytorch [51].
- Pytorch Lightning [52]
- Numpy [53].
- Torch spatio-temporal [44].
- Pandas [54,55].
- Hydra [56].

To streamline the execution of experiments, a Docker image and corresponding container [57] were created, along with scripts to facilitate their setup and use. All experimental results were obtained using a system configured as follows: Ubuntu 22.04.2 LTS operating system, an AMD Ryzen Threadripper PRO 3955WX CPU with 16 cores, an NVIDIA RTX A5000 GPU with 24 GB of memory, and 128 GB of DDR4 RAM (8x16 GB modules). Table A.13 summarizes the training and inference times achieved with this setup.